\documentclass[sigconf]{acmart}

\usepackage{booktabs} 
\usepackage{subfigure}
\usepackage{enumitem}
\usepackage{ragged2e}

\copyrightyear{2018}
\acmYear{2018} 
\setcopyright{iw3c2w3}
\acmConference[WWW '18 Companion]{The 2018 Web Conference Companion}{April 23--27, 2018}{Lyon, France}
\acmBooktitle{WWW '18 Companion: The 2018 Web Conference Companion, April 23--27, 2018, Lyon, France}
\acmPrice{}
\acmDOI{10.1145/3184558.3186584}
\acmISBN{978-1-4503-5640-4/18/04}

\begin{document}
\title{Fine-grained Video Attractiveness Prediction Using Multimodal\\  Deep Learning on a Large Real-world Dataset}

\author{
  Xinpeng Chen$^{\dagger\ast}$, Jingyuan Chen$^{\flat\ast}$, Lin Ma$^{\ddagger\natural}$, Jian Yao$^{\dagger}$, Wei Liu$^{\ddagger\natural}$, Jiebo Luo$^{\S}$, Tong Zhang$^{\ddagger}$ \\
 $^\dagger$Wuhan University, $^{\ddagger}$Tencent AI Lab, $^{\flat}$National University of Singapore,   $^{\S}$University of Rochester}

\thanks{$^\ast$ Work done while Xinpeng Chen and Jingyuan Chen were Research Interns with Tencent AI Lab. \\
$^{\natural}$Correspondence: {forest.linma@gmail.com; wliu@ee.columbia.edu.}
}


\begin{abstract}
Nowadays, billions of videos are online ready to be viewed and shared. Among an enormous volume of videos, some popular ones are widely viewed by online users while the majority attract little attention. Furthermore, within each video, different segments may attract significantly different numbers of views. This phenomenon leads to a challenging yet important problem, namely fine-grained video attractiveness prediction, which only relies on video contents to forecast video attractiveness at fine-grained levels, specifically video segments of several second length in this paper. However, one major obstacle for such a challenging problem is that no suitable benchmark dataset currently exists. To this end, we construct the first fine-grained video attractiveness dataset (FVAD), which is collected from one of the most popular video websites in the world. In total, the constructed FVAD consists of $1,019$ drama episodes with $780.6$ hours covering different categories and a wide variety of video contents. Apart from the large amount of videos, hundreds of millions of user behaviors during watching videos are also included, such as ``view counts'', ``fast-forward'', ``fast-rewind', and so on, where ``view counts'' reflects the video attractiveness while other engagements capture the interactions between the viewers and videos. First, we demonstrate that video attractiveness and different engagements present different relationships. Second, FVAD provides us an opportunity to study the fine-grained video attractiveness prediction problem.  We design different sequential models to perform video attractiveness prediction by relying solely on video contents. The sequential models exploit the multimodal relationships between visual and audio components of the video contents at different levels.
Experimental results demonstrate the effectiveness of our proposed sequential models with different visual and audio representations, the necessity of incorporating the two modalities, and the complementary behaviors of the sequential prediction models at different levels.
\end{abstract}

%
%



\maketitle

\section{Introduction}
\begin{figure}[!t]
    \centering
    \includegraphics[width=\linewidth]{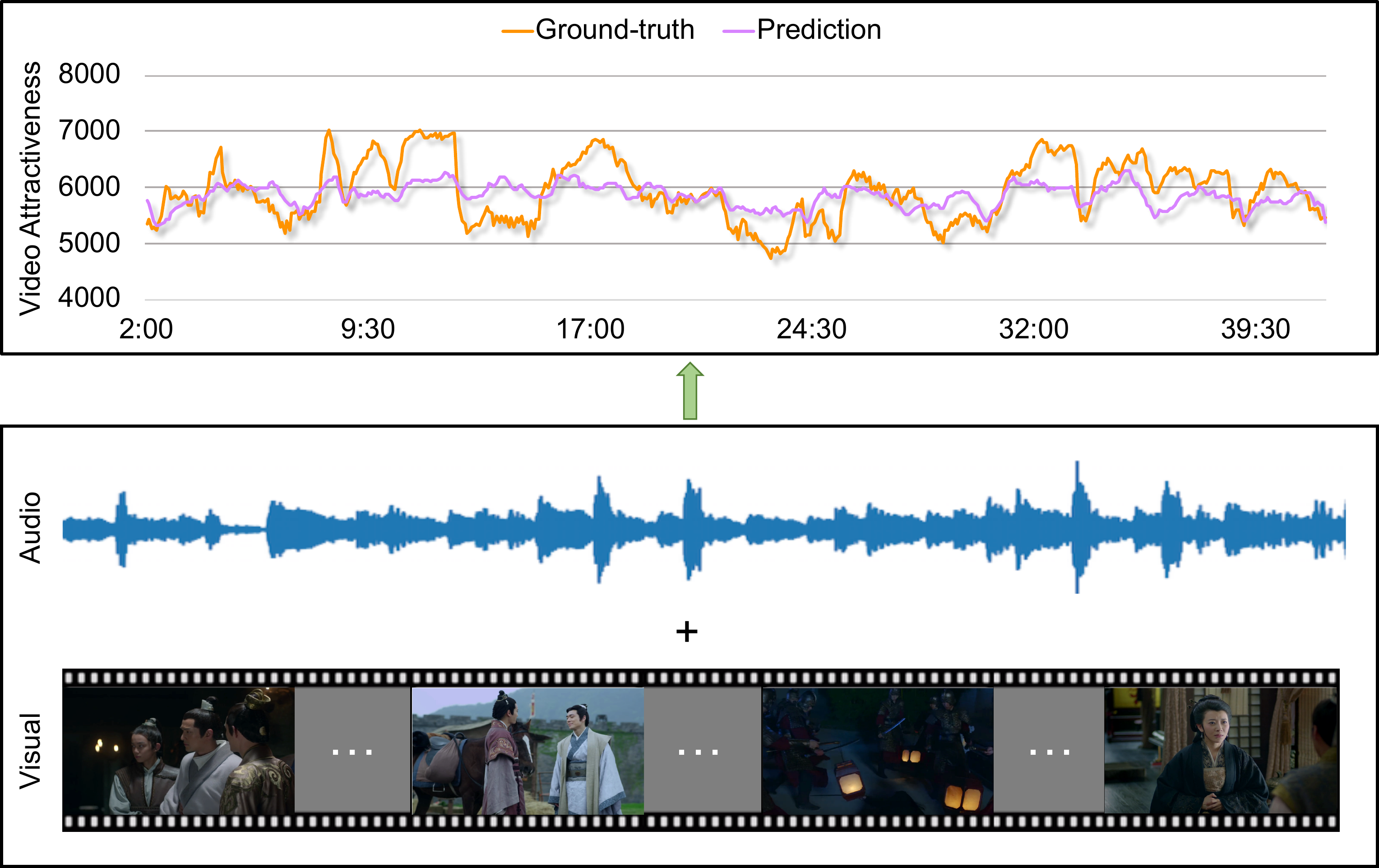}
    \vspace{-1em}
    \caption{Definition of fine-grained video attractiveness prediction.  The view counts (video attractiveness) for fine-grained video segments are shown, where the orange line denotes the ground-truth view counts based on hundreds of millions of active users. It can be observed that video attractiveness varies significantly over time. The main reason is that the contents of different video segments are of great diversity, with different visual information (\textit{e.g.}, peaceful scenery vs. fierce battle) and different audio information (\textit{e.g.}, soft background music vs. meaningful conversations). Such visual and audio contents together greatly influence the video attractiveness. Note that the purple line predicted based on both the visual and audio data using our proposed model in Sec.~\ref{sec:attractiveness} can well track the trends of the ground-truth video attractiveness.}
    \label{fig1}
\end{figure}

Today, digital videos are booming on the Internet. It is stated that traffic from online videos will constitute over 80\% of all consumer Internet traffic by 2020\footnote{\url{https://goo.gl/DrrKcn}.}. Meanwhile, due to the advance of mobile devices, millions of new videos are streaming into the Web everyday. Interestingly, among an enormous volume of videos, only a small number of them are attractive to draw a great number of viewers, while the majority receive little attention. Even within the same video, different segments present different attractiveness to the audiences with a large variance. A video or segment is considered attractive if it gains a high view count based on the statistics gathered on a large number of users. The larger the view count is, the more attractive the corresponding video or segment is. The view count directly reflects general viewers' preferences, which are thus regarded as the sole indicator of the video attractiveness within the scope of this paper.  Considering one episode from a hot TV series, as an example in Fig~\ref{fig1}~(a), the orange line indicates the view counts (attractiveness) for the short video segments, which are crawled from one of the most popular video websites. As can be seen, video attractiveness varies greatly over different video segments, where the maximum view count is more than twice of the minimum value.

Predicting the attractiveness of video segments in advance can benefit many applications, such as online marketing~\cite{DBLP:conf/mm/ChenSNWZC16} and video recommendation~\cite{DBLP:conf/sigir/ChenZ0NLC17}.
Regarding online marketing, accurate early attractiveness prediction of video segments can facilitate optimal planning of advertising campaigns and thus maximize the revenues. For video recommender systems, the proposed method provides an opportunity to recommend video segments based on their attractiveness scores.

However, predicting the video attractiveness is a very challenging task. First, the attractiveness of a video can be influenced by many external factors, such as the time that the video is posted online, the advertisement intensity in the video, and so on.  For the same category of videos, the more timely a video is delivered, the more views it will receive. Second, video attractiveness is also content-sensitive as shown in Fig.~\ref{fig1}. Therefore, in order to make reliable predictions of video attractiveness, both visual and audio contents need to be analyzed. Several existing works~\cite{SentimentFlow,Jiang,Pinto:2013:UEV:2433396.2433443,Rizoiu:2017:EHH:3038912.3052650} have explored video interestingness or popularity.  \cite{SentimentFlow,Jiang} aimed at comparing the interestingness of two videos, while \cite{Pinto:2013:UEV:2433396.2433443} relied on the historical information given by early popularity measurements. One problem is that the existing models only work on the video-level attractiveness prediction, while the fine-grained segment-level attractiveness prediction remains an open question without any attention. Another challenging problem is the lacking of large-scale real-world data. 
Recently released video datasets mostly focus on video content understanding, such as classification and captioning, specifically Sports-1M~\cite{Sports-1M}, YouTube-8M~\cite{YouTube-8M}, ActivityNet~\cite{activitynet}, UCF-101~\cite{UCF-101}, FCVID~\cite{FCVID}, and TGIF~\cite{li2016tgif}. These datasets do not incorporate any labels related to video attractiveness. In order to build reliable video attractiveness prediction systems, accurately labeled datasets are required. However, the existing video datasets for interestingness prediction~\cite{Jiang,MediaEval} are annotated by crowdsourcing. Such annotations only reflect the subjective opinions of a small number of viewers. Thus it cannot indicate the true attractiveness of the video sequence or segment.

In order to tackle the fine-grained video attractiveness prediction problem, we construct the \textbf{F}ine-grained \textbf{V}ideo \textbf{A}ttractiveness \textbf{D}ataset (FVAD), a new large-scale video benchmark for video attractiveness prediction. We collect the popular videos from one of the most popular video websites, which possesses thousands of millions of registered users. To date, FVAD contains 1,019 video episodes of 780.6 hours long in total, covering different categories and a wide variety of video contents. Moreover, the user engagements  associated with each video are also included. Besides the view counts (attractiveness), there are other 9 types of engagement indicators associated with a video sequence to record the interactions between the viewers and videos, as illustrated in Fig.~\ref{fig2}. We summarize our contributions as follows:
\begin{itemize}
  \item We build the largest real-world dataset FVAD for dealing with the task of fine-grained video attractiveness prediction. The video sequences and their associated ``labels'' in the form of view count, as well as the viewers' engagements with videos are provided. The relationships between video attractiveness and engagements are examined and studied. 
  \item Several sequential models for exploiting the relationships between visual and audio components for fine-grained video attractiveness prediction are proposed. Experimental results demonstrate the effectiveness of our proposed models and the necessity of jointly considering both the visual and audio modalities.  
\end{itemize}

\begin{figure*}[t]
    \centering
    \includegraphics[width=\linewidth]{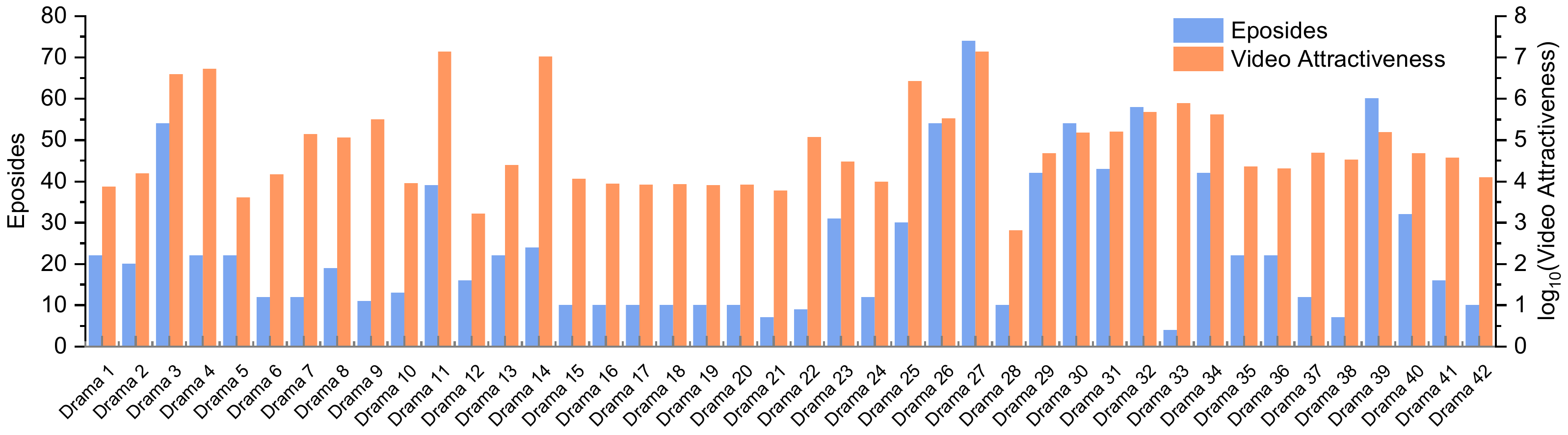}
    \vspace{-1em}
    \caption{The statistics of 42 kinds of dramas in our constructed FVAD. The blue bars indicate the number of videos per TV series, while the orange ones indicate the video attractiveness (view counts) in the $\mathbf{Log}_{10}$ domain.}
    \label{fig_video_statistics}
\end{figure*}

\section{Related Work}
\subsection{Video Datasets}
Video datasets have played a critical role in advancing computer vision algorithms for video understanding. Several well labeled small-scale datasets, such as XM2VTS~\cite{Messer-avbpa99}, KTH~\cite{KTH}, Hollywood-2~\cite{Hollywood2}, Weizmann~\cite{weizmann}, UCF101~\cite{UCF-101}, THUMOS'15~\cite{THUMOS15}, HMDB~\cite{HMDB}, and ActivityNet~\cite{activitynet}, provide benchmarks for face recognition~\cite{liu2006spatio}, human action recognition~\cite{jiang2012trajectory} and activity understanding. There are also other video datasets focusing on visual content recognition, video captioning, and so on, such as FCVID~\cite{FCVID} and TGIF~\cite{li2016tgif}. In order to make a full exploitation on the video content understanding, super large video datasets have been recently constructed. Sports-1M~\cite{Sports-1M} is a dataset for sports video classification with 1 million videos. YFCC'14~\cite{YFCC} is a large multimedia dataset including about 0.8 million videos. The recent YouTube-8M~\cite{YouTube-8M} is so far the  largest dataset for multi-label video classification, consisting of about 8 million videos. However, it is prohibitively expensive and time consuming to obtain a massive amount of well-labeled data. Therefore, these datasets inevitably introduce label noise when the labels are produced automatically. The most important thing is that all these datasets focus on understanding video contents, without touching on the video attractiveness task. MediaEval~\cite{MediaEval} is the only known public dataset, which is closely  related to our work. It is used for predicting the interesting frames in movie trailers. However, MediaEval is a small dataset that only consists of 52 trailers for training and 26 trailers for testing. In addition, the interesting frames in MediaEval are labeled by a small number of subjects, which is not consistent with the real-life situation of massive diverse audiences. 

\subsection{Video Attractiveness Prediction}
A thread of work for  predicting video interestingness or popularity is related to our proposed video attractiveness prediction. In~\cite{Liu2009}, where Flickr images are used to measure the interestingness of video frames. Flickr images were assumed to be mostly interesting compared with many video frames since the former are generally well-composed and selected for sharing. A video frame is considered interesting if it matches (using image local features) with a large number of Flickr images. In~\cite{SentimentFlow}, after extracting and combining the static and temporal features using kernel tricks, a relative score is predicted to determine which video is more interesting than the other using ranking SVM given a pair of videos. In~\cite{Jiang}, two datasets are collected based on interestingness ranking from Flickr and YouTube, and the interestingness of a video is predicted in the same way as~\cite{SentimentFlow}. In~\cite{Pinto:2013:UEV:2433396.2433443}, the historical information given by early popularity measurements is used for video popularity prediction. A Hawkes intensity process is proposed to explain the complex popularity history of each video according to its type of content, network of diffusion, and sensitive to promotion~\cite{Rizoiu:2017:EHH:3038912.3052650}. Different from \cite{Liu2009,SentimentFlow,Jiang}, video contents are not explicitly used for video popularity prediction~\cite{Pinto:2013:UEV:2433396.2433443,Rizoiu:2017:EHH:3038912.3052650}.

Our work is fundamentally different from the previous works. First, large-scale real-world user behaviors on one of the most popular video websites, are crawled to construct the proposed FVAD. Second,we aim to predict the fine-grained actual video attractiveness (view counts), compared with the video-level interestingness~\cite{Liu2009,SentimentFlow,Jiang} and popularity~\cite{Pinto:2013:UEV:2433396.2433443,Rizoiu:2017:EHH:3038912.3052650}. Third, we develop different sequential multimodal models to jointly learn the relationships between visual and audio components for the video attractiveness prediction. To the best of our knowledge, there is no existing work to handle and study the fine-grained video attractiveness prediction problem.

\begin{figure*}[!t]
    \centering
    \includegraphics[width=\linewidth]{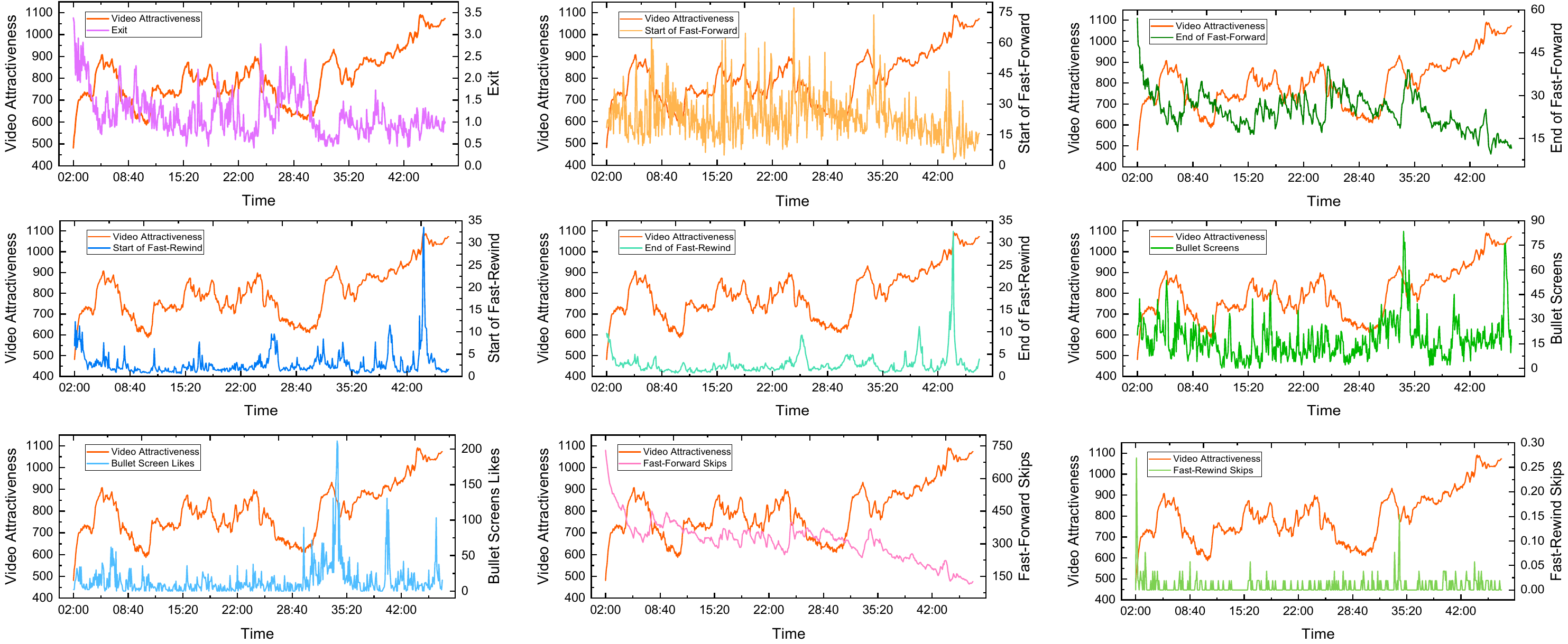}
    \vspace{-2em}
    \caption{The other 9 types of viewers' engagements with the video sequences while watching. The view counts are also presented together with each engagement. It can be observed that the view counts present different correlations to these 9 types of viewers' engagements. From top-left to bottom-right: 1) Exit: the number of viewers exiting the show, 2) Start of Fast-Forward (FF): the number of viewers beginning FF,  3) End of Fast-Forward: the number of viewers stopping FF,  4) Start of Fast-Rewind (FR): the number of viewers beginning FR, 5) End of Fast-Rewind: the number of viewers stopping FR, 6) Bullet Screens: the number of bullet screens sent by viewers, 7) Bullet Screen Likes: the number of  bullet screen likes of the viewers, 8) Fast-Forward Skips: the number of skip times during FF, and  9) Fast-Rewind Skips: the number of skip times during FR.}
    \label{fig2}
\end{figure*}

\section{FVAD Construction}
This section elaborates on the FVAD dataset construction, covering  the
video collecting strategy, the video attractiveness and engagements, and the analysis of their relationships.

\subsection{Video Collection}
To construct a representative dataset which contains video segments with diverse attractiveness degrees, video contents should cover different categories and present a broad range of diversities. We manually select a set of popular TV serials from the website. For different episodes and fragments within each episode, as the story develops, it is obvious that the attractiveness degree goes upward and downward.  As shown in Fig. \ref{fig1}, the video contents, including the visual and audio components, significantly affect the video attractiveness presenting diverse view counts. For our FVAD dataset, we collected $1,019$ episodes with a total duration of $780.6$ hours long. The number of episodes with respect to each TV series is illustrated by the blue bars in Fig.~$\ref{fig_video_statistics}$. The average duration of all the episodes in FVAD is $45$ minutes. Moreover, all the episodes were downloaded in high quality with the resolution of $640\times480$.

\subsection{Video Attractiveness}

\begin{figure}[t]
    \centering
    \includegraphics[scale=0.39]{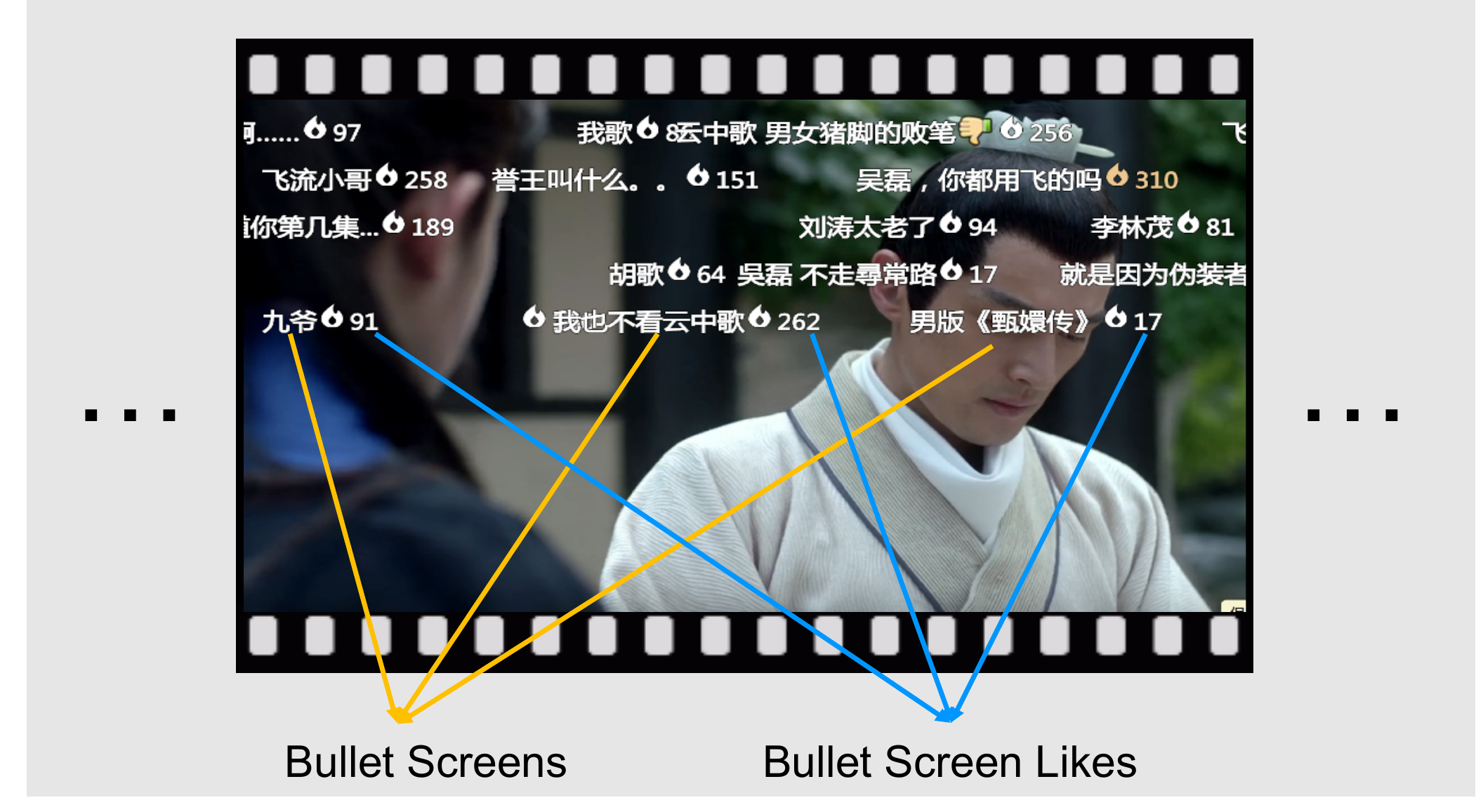}
    \vspace{-1em}
    \caption{ A simple example of bullet screens. Different users may
express real-time opinions directly upon their interested
frames.}
    \label{meta-data_explanation}
    \vspace{-10pt}
\end{figure}

In this paper, we focus on the fine-grained video attractiveness. Therefore, we need to collect the attractiveness indicators of the fine-grained video fragments. 
As aforementioned, the attractiveness degree for each video fragment is quantified by the total number of views. As shown in~\cite{DBLP:journals/cacm/SzaboH10}, visual media tends to receive views over some period
of time. To normalize this effect, we divide the number
of views by the duration from the upload date of the given
episode to the collection date, which is 30th November, 2017. The orange bar in Fig. \ref{fig_video_statistics} illustrates the total video attractiveness of the TV series by summing all the view counts from all the episodes in each season. In order to make a better visualization, the attractiveness value is displayed in the $\mathbf{Log}_{10}$ domain. It can be observed that the video attractiveness varies significantly among different TV series. Even for the same TV series, different seasons present different attractiveness.

\subsection{Video Engagements}
In addition to video views, we also collected $9$ user engagement-related indicators {regarding each video fragment}, namely Exit, Start of Fast-Forward, End of Fast-Forward, Start of Fast-Rewind, End of Fast-Rewind, Fast-Forward Skips, Fast-Rewind Skips, Bullet Screens, and Bullet Screen Likes. The first 7 engagements are the natural user behaviors during the watching process, while the last two engagements, namely the Bullet Screens and Bullet Screen likes, involve deep interactions between viewers and videos. 

Bullet Screens, also named as  time-synchronized comments and first introduced in~\cite{DBLP:conf/kdd/WuZTH014}, allow users to express opinions directly on the frames of interest in a real-time manner. Intuitively, the user behaviors of commenting on a frame can be regarded as implicit feedback reflecting the frame-level preference, while the image features of the reviewed frame and the textual features in the posted comments can further help model the fine-grained preference from different perspectives. Fig.~\ref{meta-data_explanation} shows a simple example of bullet screen. As can be seen, different users may express real-time opinions directly upon their interested frames. The number after each bullet screen in Fig.~\ref{meta-data_explanation} indicates the total number of likes received by the corresponding bullet screen from the audience. The comment words from Bullet Screens can more accurately express the viewers' preferences and opinions. However, for this paper, we only collect the numbers of the Bullet Screens as well as their associated number of likes.

Fig.~$\ref{fig2}$ illustrates  the $9$ different engagement indicators as well as the video attractiveness of one episode. It is noticed that the distributions of these different engagements  are different. Each of them measures one aspect of users' engagement behaviors. These engagement characters intuitively correlate with the video attractiveness indicator (view counts). For example, high Fast-Forward Skips values always correspond to low attractiveness, while high Start of Fast-Rewind values correspond to high attractiveness.

\subsection{Relationships between Video Attractiveness and Engagements}
To evaluate the above correlations quantitatively, three kinds of coefficients, including Pearson correlation coefficient~(PCC), cosine similarity~(CS), and Spearman rank-order correlation coefficient~(SRCC), are used to measure the strength and direction of the association between each engagement indicator and attractiveness. The correlations are provided in Table~\ref{meta-data_correlation}. It is demonstrated that different engagement indicators present different correlations with the attractiveness, where some present positive correlations while others present negative correlations. It is not surprising that the {Start of Fast-Forward} and {Fast-Forward Skips} present the largest positive and negative correlations, respectively. However, the indicator {Bullet Screens} shows negative correlations with video views. One possible reason is that the actual commented frame should be the one corresponding to the time when the user began to type the bullet screen, rather than the frame when the bullet screen was posted out. Therefore, the main reason is that the data about bullet screens is not well aligned. Another possible reason is that most bullet screens are complaints about the stories, therefore being not able to represent the attractiveness of the video. It is noted that both {Bullet Screen Likes} and {Fast-Rewind Skips} show less correlations with video views. One possible reason is that the value of each indicator is relatively small, which thereby cannot reflect statistical regularities.

\begin{table}
  \begin{center}
  \small
  \caption{The correlations between video attractiveness and different engagement indicators, in terms of Pearson correlation coefficient~(PCC), cosine similarity~(CS), and spearman's rank correlation coefficient~(SRCC).}
  \label{meta-data_correlation}
  \vspace{-1em}
  \begin{tabular}{|l|c|c|c|}
    \hline
    Indicator Name & PCC & CS & SRCC \\
    \hline\hline
    Exit & -0.149 & -0.148 & -0.210 \\
    Start of Fast-Forward & -0.117 & -0.117 & -0.200 \\
    End of Fast-Forward & -0.537 & -0.536 & -0.522 \\
    Start of Fast-Rewind & 0.327 & 0.327 & 0.368 \\
    End of Fast-Rewind & 0.227 & 0.227 & 0.256 \\
    Bullet Screens & -0.139 & -0.139 & -0.191 \\
    Bullet Screen Likes & 0.027 & 0.027 & -0.020 \\
    Fast-Forward Skips & -0.351 & -0.350 & -0.315 \\
    Fast-Rewind Skips & 0.022 & 0.022 & 0.013 \\
    \hline
  \end{tabular}
  \end{center}
\end{table}

\section{Video Attractiveness Prediction Using Deep Learning on Large Datasets}
\label{sec:attractiveness}
Video attractiveness prediction is a very challenging task, which may involve many external factors. 
For example, social influence is an important external factor, which makes a great impact on the number of views. In the Western world, the drama series such as \textit{The Big Bang Theory} have a huge number of fans, which are of high attractiveness. However, for Chinese viewers, \textit{The Big Bang Theory} are less attractiveness than some reality shows, such as \textit{I Am a Singer}. In the constructed FVAD, since user profile data is not available, we cannot track users' culture backgrounds or consider other social-related factors. Another important external factor is the director and starring list of the corresponding TV series. Specifically, a strong cast always boosts the base attractiveness of the whole series. For example, some dramas such as \emph{Empresses in the Palace} with many famous stars attract billions of views.

Besides different external factors, video contents play the most important role in the task of video attractiveness prediction. In this paper, we aim at discovering the relationships between video contents and video attractiveness. Even further, we would like to make the prediction on the video attractiveness based solely on the video contents. Therefore, we need to first eliminate the effects of external factors. We use one simple method, namely the standardization, on the attractiveness as well as the other 9 engagement indicators. With such normalization, we can obtain the video relative attractiveness, which is regarded to be determined by the video contents only, specifically the visual and audio components. In the following, we will employ the normalized video attractiveness to perform the video attractiveness prediction. 

\begin{figure*}[t]
    \centering
    \subfigure{\includegraphics[scale=0.31]{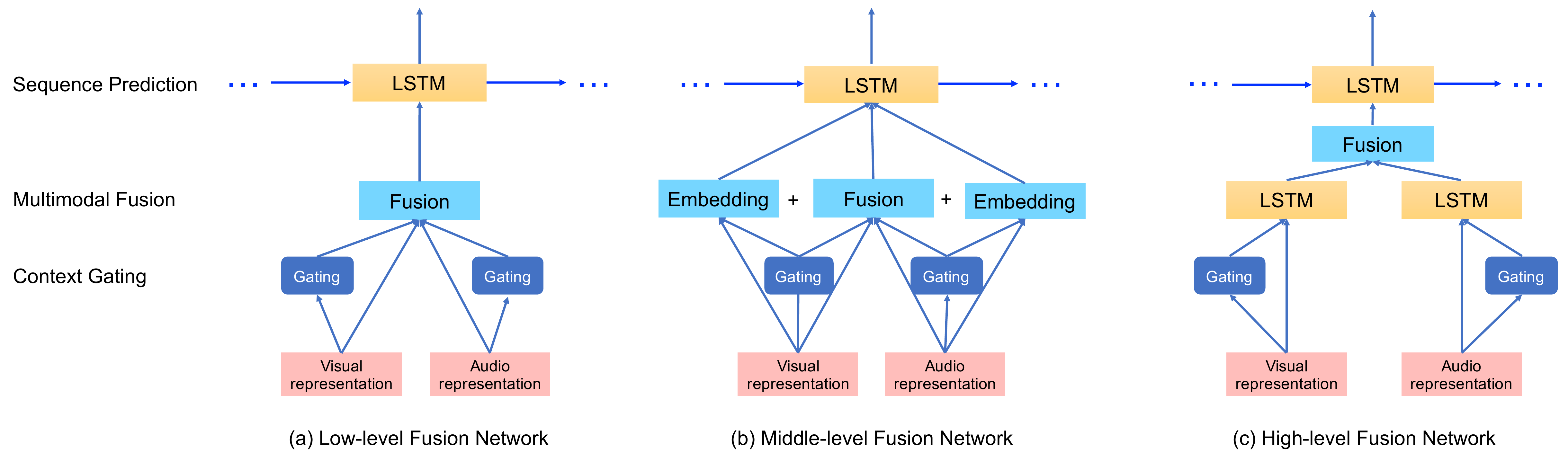}}
    \vspace{-1.5em}
    \caption{An overview of our video attractiveness prediction framework. Context gating is first applied to the visual and audio representations to enrich their corresponding representations. Based on the gated representations, different multimodal fusion strategies performed at different levels are used to exploit the relationships between visual and audio components. Finally, LSTM acts as the prediction layer to make the attractiveness prediction.}
    \label{predict_framework}
    \vspace{-10pt}
\end{figure*}

\subsection{Video Representation}
\label{sec:video_rep}
To comprehensively understand video contents, we extract both visual and audio representations.

\textbf{Visual representation.} 
Recently developed convolutional neural networks (CNNs), such as VGG~\cite{vgg}, Inception-X~\cite{InceptionV1,InceptionV2,InceptionV3,InceptionV4} and ResNet~\cite{ResNet}, are usually utilized to generate global representations of images. Relying on these CNNs, we decode each video with FFmpeg, select  1 frame per second, feed each visual frame into a CNN model, and fetch the hidden states before the classification layer as  the visual feature. Specifically, to exploit the capacity of different kinds of CNN models, we experiment a variety of CNNs, namely VGG-16, VGG-19, ResNet-152, Inception-X, and the recently developed model NasNet~\cite{nasnet}.

\textbf{Audio representation.}
For the acoustic modality, mel-frequency cepstral coefficient~(MFCC)~\cite{mfcc1980} is widely used in many audio-related tasks~\cite{tzanetakis2002musical,gowdy2000mel}. In this paper, MFCC feature is also used for audio representation. Specifically, for a given audio file, the length of the sampling window is set to $25$ milliseconds and meanwhile the step between successive windows is set to $10$ milliseconds. In this way, there will be $100$ MFCC features per second. To reduce the feature dimension, we take the average of the MFCC feature every second. Since there are two channels in the audio file, we first extract the MFCC features for each channel and then concatenate them together. As a result, the dimension of the MFCC feature for a given audio signal is $T \times 26$, where $T$ is the length of a audio signal. In addition to MFCC feature, we also use NSynth~\cite{engel2017neural} to encode the audio signals. NSynth is a recently developed WaveNet-style~\cite{wavenet} auto-encoder model. Concretely, we take audio fragment every $5$ seconds as input into NSynth and get the output of the encoder as the audio representation.

\subsection{Proposed Multimodal Deep Learning Models}
Our proposed multimodal deep learning model for video attractiveness prediction consists of three layers, namely the context gating layer, the multimodal fusion layer, and the sequential prediction layer.

\textbf{Context gating layer.}
In order to further enrich the representative properties of the visual and audio features, context gating is used, which is  shown to be beneficial to video representation learning~\cite{miech2017learnable}. Context gating is formulated as:
$$\hat{X} = \sigma (WX + b) \odot X,$$
where $X$ is the input feature vector, which can either be visual or audio representation. $\sigma$ is the element-wise sigmoid activation function. $\odot$ denotes the element-wise multiplication. $\hat{X}$ is the gated representation. It can be observed that  context gating  acts like a sentinel, which can adaptively decide which part of the input feature is useful. Moreover, with multiplication, the original representation $X$ and the transformed representation $\sigma(Wx+b)$ are nonlinearly fused together, thus enhancing and enriching their representative abilities.

\textbf{Multimodal fusion layer.}
Video contents consist of both visual and audio information, which are complementary to each other for the video representation learning~\cite{YouTube-8M}. Therefore, in this paper, we propose several multimodal fusion models to exploit the relationships between the gated visual and audio features to yield the final video representation. As illustrated in Fig.~\ref{predict_framework}, three different multimodal fusion layers performed at different levels are proposed to yield the video representation for the final attractiveness prediction. 

\begin{table*}[t]
  \small
  \begin{center}
  \caption{Performance comparisons of our proposed multimodal deep learning models with different visual and audio representations, as well as their combinations. The best performance  (except LSTM-EGG) for each metric entry is highlighted in boldface.}
  \label{performance_results}
  \vspace{-1.2em}
  \begin{tabular}{|l|c|c|c|c|}
    \hline
    Model Name & SRCC~($\rho$) & MAE & RMSE & RMSLE \\
    \hline\hline
    LSTM-EGG & 0.795 & 0.381 & 0.499 & 0.039 \\
    \hline\hline
    LSTM-AUD-MFCC~\cite{mfcc1980}  & 0.210 & 0.600 & 0.775 & 0.076 \\
    LSTM-AUD-NSynth~\cite{wavenet,engel2017neural}  & 0.213 & 0.606 & 0.802 & 0.082 \\
    \hline\hline
    LSTM-VIS-VGG-16~\cite{vgg} & 0.323 & 0.572 & 0.726 & 0.069 \\
    LSTM-VIS-VGG-19~\cite{vgg} & 0.322 & 0.569 & 0.725 & 0.067 \\
    LSTM-VIS-ResNet-152~\cite{ResNet} & 0.241 & 0.602 & 0.773 & 0.075 \\
    LSTM-VIS-NasNet-large~\cite{nasnet} & 0.359 & 0.570 & 0.724 & 0.069 \\
    LSTM-VIS-Inception-V1~\cite{InceptionV1} & 0.336 & 0.570 & 0.719 & 0.066 \\
    LSTM-VIS-Inception-V2~\cite{InceptionV2} & 0.337 & 0.569 & 0.724 & 0.067 \\
    LSTM-VIS-Inception-V3~\cite{InceptionV3} & 0.335 & 0.571 & 0.725 & 0.068 \\
    LSTM-VIS-Inception-V4~\cite{InceptionV4} & 0.365 & 0.567 & 0.713 & 0.067 \\
    \hline\hline
    Low-level fusion~(Inception-V4+MFCC) & 0.313 & 0.580 & 0.740 & 0.070 \\
    Low-level fusion~(Inception-V4+NSynth) & 0.243 & 0.601 & 0.793 & 0.079 \\
    \hline\hline
    Middle-level fusion~(Inception-V4+MFCC) & 0.330 & 0.575 & 0.731 & 0.069 \\
    Middle-level fusion~(Inception-V4+NSynth) & 0.318 & 0.573 & 0.733 & 0.070 \\
    \hline\hline
    High-level fusion~(Inception-V4+MFCC) & 0.387 & 0.562 & 0.708 & 0.066 \\
    High-level fusion~(Inception-V4+NSynth) & 0.371 & 0.551 & 0.698 & 0.063 \\
    \hline\hline
    Ensemble of high, middle and low level fusion~(Inception-V4+MFCC) & \textbf{0.401} & 0.554 & 0.699 & 0.065 \\
    Ensemble of high, middle and low level fusion~(Inception-V4+NSynth) & 0.393 & \textbf{0.544} & \textbf{0.690} & \textbf{0.062} \\
    \hline
  \end{tabular}
  \end{center}
  \vspace{-10pt}
\end{table*}

{\textit{Low-level fusion}}.
Fig.~\ref{predict_framework}~(a) illustrates the low-level fusion layer. Specifically, we directly concatenate the visual and audio features after the aforementioned context gating layer and project them into a common space with a single embedding layer. As such, the low-level fusion strategy allows the visual and audio features to be fused at low levels. However, the contributions of visual and audio modalities are not equal. Normally, visual components will present more semantic information than audio. Simply concatenating them together may make the audio information be concealed by the visual part. 

{\textit{Middle-level fusion}}.
To tackle the information concealment problem, we propose a middle-level fusion layer to learn the comprehensive representations from the two modalities. The architecture is shown in Fig.~\ref{predict_framework}~(b).  Specifically, we transform the gated visual and audio features with non-linear operations into three independent embeddings: the visual  embedding, the audio embedding, and the joint embedding. The joint embedding captures the common semantic meanings between visual and audio modalities, while the visual and audio embeddings capture the corresponding independent semantic meanings.

{\textit{High-level fusion}}.
Furthermore, to fully exploit the temporal relations among the representations at every time step, we propose a more effective fusion method which is termed as the high-level fusion layer. As illustrated in Fig.~\ref{predict_framework}~(c), we take two individual long short-term memory (LSTM) networks to encode the features of visual and audio data into the higher-order representations, which are further fused together as the video representation for the attractiveness prediction. With two different yet dependent LSTMs employed to learn the complicated behaviors within each individual modality, the semantic meanings carried by visual and audio components are extensively discovered, which is expected to benefit the final video attractiveness prediction.

\textbf{Sequential prediction layer.}
After we obtain the multimodal embedding with both visual and audio components considered, we use a sequential prediction network to estimate the video attractiveness. More specifically, we take the output of the multimodal fusion layer $x_t$ at $t$-th time step as input of another LSTM  for prediction. We formulate the prediction process as follows:
\begin{align}
h_t = \textrm{LSTM}(x_t, h_{t-1}).
\end{align}
The LSTM transition process is formulated as follows:
\begin{equation} 
\label{lstm_linear}
\begin{split}
  \begin{pmatrix} i_t \\ f_t \\ o_t \\ g_t \end{pmatrix} &=
  \begin{pmatrix} \sigma \\ \sigma \\ \sigma \\ \tanh \end{pmatrix}
  \mathbf{T}
  \begin{pmatrix} x_t \\ h_{t-1} \end{pmatrix},\\
  c_t &= f_t \odot c_{t-1} + i_t \odot g_t, \\
  h_t &= o_t \odot \tanh(c_t), \\
  y^{\prime} &= W_{o}(h_t),
\end{split}
\end{equation}
\noindent where $i_t$, $f_t$, $o_t$, $c_t$, $h_t$, and $\sigma$ are input gate, forget gate, output gate, memory cell, hidden state, and sigmoid function, respectively. $\mathbf{T}$ is a linear transformation matrix. $\odot$ represents an element-wise product operator. The hidden state $h_t$ is used to predict a value $y^{\prime}$ as video attractiveness at fine-grained levels through a linear transformation layer $W_{o}$.

\subsection{Training}
Mean squared error~(MSE) is a widely used as the objective function in sequence prediction tasks, which can be formulated as follows:
\begin{equation}
\mathcal{L}_{\text{MSE}} = \sum_{i=1}^{T}{(y_{i}^{\prime} - y_{i})^2}.
\end{equation}
$y_{i}^{\prime}$ is the attractiveness value predicted by our model. $y_{i}$ is the ground truth attractiveness (view counts). $T$ is the fragment length of the video clip. Then we can use gradient descent methods to train the whole model in an end-to-end fashion.

\section{Experiments}
\label{sec:experiments}

In this section, we first introduce the experiment settings, including the data processing, evaluation metrics, baselines, as well as our implementation details. Afterward, we will illustrate and discuss the experimental results. 

\subsection{Experimental Settings}
\textbf{Data processing.}
To keep the diversity of training samples, for episodes in each category\footnote{A season of TV series can be seen as a category in this scenario.}, we use $70$\% for training, $20$\% for testing and $10$\% for validation. Recall that the average duration of videos in FVAD is $45$ minutes, which is difficult for LSTM to model such long video sequence due to the capacity limitations of LSTM. Therefore, we divide each video in the training set into a series of non-overlapping video clips with the length of $5$ minutes. However, during the testing phase of our model, we take the video as a whole into the prediction model without any partitioning. 

\textbf{Evaluation metrics.}
To evaluate the performance of fine-grained video attractiveness prediction, we adopted mean absolute error (MAE), root mean square error~(RMSE) and root mean squared logarithmic error~(RMSLE). Besides, as in~\cite{khosla2014makes}, we adopt Spearman rank-order correlation coefficient~(SRCC) to evaluate the correlation between the video attractiveness predicted by our model and the true values. According to the definitions, larger SRCC value and smaller MAE, RMSE, and RMSLE values indicate more accurate predictions, demonstrating a better performance. 

\textbf{Baselines.}
The framework of our baseline models is similar to the model illustrated in Fig.~\ref{predict_framework}~(a). The only difference is that the baseline model only takes one kind of feature as input. More specifically, given any types of representation $X$, we first transform $X$ into an embedding vector with dimension size of $512$. Then the embedding vector is input into the sequence prediction layer to estimate the video attractiveness. In our experiments, \textit{LSTM-EGG} represents the model which predicts the attractiveness with $9$ engagement indicators. \textit{LSTM-AUD-$\ast$} and \textit{LSTM-VIS-$\ast$} are the baseline models which only take the audio and the visual representations as input, respectively. 

\textbf{Implementation details.}
In this paper, the hidden unit size of LSTM are all set to 512. We train the model with the adam optimizer by a fixed learning rate $5 \times 10^{-4}$. The batch size is set as 16. And the training procedure is terminated with early stopping strategy when value of ($3 \times \text{SRCC} - \text{MAE} - \text{RMSE} - \text{RMSLE}$) reaches the maximum value on the validation set.

\subsection{Results and Discussions}
The experimental results are illustrated in Table~$\ref{performance_results}$. Different video and audio representations, as well as their variant combinations, are used to perform the visual attractiveness prediction. 

Recall that in Section $3$ we verified that there indeed exists correlations between video attractiveness and other user engagement indicators. To investigate the combined effect of all engagement indicators, we show the performance of LSTM-EGG. We observed that LSTM-EGG obtains the best result which indicates that users' engagement behaviors as a whole shows a strong correlation with video attractiveness (view counts). This also validates that the features developed from engagement domain are much discriminative, even though they are of low-dimension. However, such features are not available for practical applications. That is also the main reason why we resort to the content features, specifically the visual and audio contents, for video attractiveness prediction.

Through the comparison among LSTM-AUD-$\ast$, LSTM-VIS-$\ast$ and different fusion methods, it is observed that visual features are more useful than audio features for video attractiveness prediction. Moreover, by incorporating more modalities, better performances can be obtained. This implies the complementary relationships rather than mutual conflicting relationships between the visual and audio modalities. To further examine the discriminative properties of the audio and visual features, we conduct experiments over different kinds of features using the proposed model. The general trend is that the more powerful the visual or audio features, the better performance it obtained. Specifically, the visual features in the form of NasNet and Inception-X are more powerful than those of VGG.

It is obvious that high-level fusion performs much better than low-level fusion methods. Regarding the low-level fusion, features extracted from various sources may not fall into the same common space. Simply concatenating all features actually brings in a certain amount of noise and ambiguity. Besides, low-level fusion may lead to the curse of dimensionality since the final feature vector would be of very high dimension. High-level fusion methods introduce two separate LSTMs to well capture the semantic meanings of the visual and audio content, respectively, which thus make a more comprehensive understanding of video contents. Additionally, the ensemble results among all levels of fusion achieve the best performance, which demonstrating that ensembling different level fusion models can comprehensively exploit the video content for attractiveness prediction.

\section{Conclusions}
In this paper, we built to date the largest benchmark dataset, dubbed FVAD, for tackling the emerging fine-grained video attractiveness prediction problem. The dataset was collected from a real-world video website. Based on FVAD, we first investigated the correlations between video attractiveness and nine user engagement behaviors. In addition, we extracted a rich set of attractiveness oriented features to characterize the videos from both visual and audio perspectives. Moreover, three multimodal deep learning models were proposed to predict the fine-grained fragment-level attractiveness relying solely on the video contents. Different levels of multimodal fusion strategies were explored to model the interactions between visual and audio modalities. Experimental results demonstrate the effectiveness of the proposed models and the necessity of incorporating both the visual and audio modalities.

\bibliographystyle{ACM-Reference-Format}
\bibliography{sifconf_bibliography}


\begin{thebibliography}{39}


\ifx \showCODEN    \undefined \def \showCODEN     #1{\unskip}     \fi
\ifx \showDOI      \undefined \def \showDOI       #1{#1}\fi
\ifx \showISBNx    \undefined \def \showISBNx     #1{\unskip}     \fi
\ifx \showISBNxiii \undefined \def \showISBNxiii  #1{\unskip}     \fi
\ifx \showISSN     \undefined \def \showISSN      #1{\unskip}     \fi
\ifx \showLCCN     \undefined \def \showLCCN      #1{\unskip}     \fi
\ifx \shownote     \undefined \def \shownote      #1{#1}          \fi
\ifx \showarticletitle \undefined \def \showarticletitle #1{#1}   \fi
\ifx \showURL      \undefined \def \showURL       {\relax}        \fi
\providecommand\bibfield[2]{#2}
\providecommand\bibinfo[2]{#2}
\providecommand\natexlab[1]{#1}
\providecommand\showeprint[2][]{arXiv:#2}

\bibitem[\protect\citeauthoryear{Abu-El-Haija, Kothari, Lee, Natsev, Toderici,
  Varadarajan, and Vijayanarasimhan}{Abu-El-Haija et~al\mbox{.}}{2016}]%
        {YouTube-8M}
\bibfield{author}{\bibinfo{person}{Sami Abu-El-Haija}, \bibinfo{person}{Nisarg
  Kothari}, \bibinfo{person}{Joonseok Lee}, \bibinfo{person}{Apostol~(Paul)
  Natsev}, \bibinfo{person}{George Toderici}, \bibinfo{person}{Balakrishnan
  Varadarajan}, {and} \bibinfo{person}{Sudheendra Vijayanarasimhan}.}
  \bibinfo{year}{2016}\natexlab{}.
\newblock \showarticletitle{YouTube-8M: A Large-Scale Video Classification
  Benchmark}. In \bibinfo{booktitle}{\emph{arXiv:1609.08675}}.
\newblock


\bibitem[\protect\citeauthoryear{Blank, Gorelick, Shechtman, Irani, and
  Basri}{Blank et~al\mbox{.}}{2005}]%
        {weizmann}
\bibfield{author}{\bibinfo{person}{Moshe Blank}, \bibinfo{person}{Lena
  Gorelick}, \bibinfo{person}{Eli Shechtman}, \bibinfo{person}{Michal Irani},
  {and} \bibinfo{person}{Ronen Basri}.} \bibinfo{year}{2005}\natexlab{}.
\newblock \showarticletitle{Actions as Space-Time Shapes}. In
  \bibinfo{booktitle}{\emph{ICCV}}.
\newblock


\bibitem[\protect\citeauthoryear{Chen, Song, Nie, Wang, Zhang, and Chua}{Chen
  et~al\mbox{.}}{2016}]%
        {DBLP:conf/mm/ChenSNWZC16}
\bibfield{author}{\bibinfo{person}{Jingyuan Chen}, \bibinfo{person}{Xuemeng
  Song}, \bibinfo{person}{Liqiang Nie}, \bibinfo{person}{Xiang Wang},
  \bibinfo{person}{Hanwang Zhang}, {and} \bibinfo{person}{Tat-Seng Chua}.}
  \bibinfo{year}{2016}\natexlab{}.
\newblock \showarticletitle{Micro Tells Macro: Predicting the Popularity of
  Micro-Videos via a Transductive Model}. In \bibinfo{booktitle}{\emph{ACM
  Multimedia}}.
\newblock


\bibitem[\protect\citeauthoryear{Chen, Zhang, He, Nie, Liu, and Chua}{Chen
  et~al\mbox{.}}{2017}]%
        {DBLP:conf/sigir/ChenZ0NLC17}
\bibfield{author}{\bibinfo{person}{Jingyuan Chen}, \bibinfo{person}{Hanwang
  Zhang}, \bibinfo{person}{Xiangnan He}, \bibinfo{person}{Liqiang Nie},
  \bibinfo{person}{Wei Liu}, {and} \bibinfo{person}{Tat-Seng Chua}.}
  \bibinfo{year}{2017}\natexlab{}.
\newblock \showarticletitle{Attentive Collaborative Filtering: Multimedia
  Recommendation with Item and Component-Level Attention}. In
  \bibinfo{booktitle}{\emph{SIGIR}}.
\newblock


\bibitem[\protect\citeauthoryear{Davis and Mermelstein}{Davis and
  Mermelstein}{1980}]%
        {mfcc1980}
\bibfield{author}{\bibinfo{person}{Steven Davis} {and} \bibinfo{person}{Paul
  Mermelstein}.} \bibinfo{year}{1980}\natexlab{}.
\newblock \showarticletitle{Comparison of parametric representations for
  monosyllabic word recognition in continuously spoken sentences}.
\newblock \bibinfo{journal}{\emph{IEEE transactions on acoustics, speech, and
  signal processing}} (\bibinfo{year}{1980}).
\newblock


\bibitem[\protect\citeauthoryear{Engel, Resnick, Roberts, Dieleman, Eck,
  Simonyan, and Norouzi}{Engel et~al\mbox{.}}{2017}]%
        {engel2017neural}
\bibfield{author}{\bibinfo{person}{Jesse Engel}, \bibinfo{person}{Cinjon
  Resnick}, \bibinfo{person}{Adam Roberts}, \bibinfo{person}{Sander Dieleman},
  \bibinfo{person}{Douglas Eck}, \bibinfo{person}{Karen Simonyan}, {and}
  \bibinfo{person}{Mohammad Norouzi}.} \bibinfo{year}{2017}\natexlab{}.
\newblock \showarticletitle{Neural Audio Synthesis of Musical Notes with
  WaveNet Autoencoders}.
\newblock \bibinfo{journal}{\emph{arXiv preprint arXiv:1704.01279}}
  (\bibinfo{year}{2017}).
\newblock


\bibitem[\protect\citeauthoryear{Gowdy and Tufekci}{Gowdy and Tufekci}{2000}]%
        {gowdy2000mel}
\bibfield{author}{\bibinfo{person}{John~N. Gowdy} {and}
  \bibinfo{person}{Zekeriya Tufekci}.} \bibinfo{year}{2000}\natexlab{}.
\newblock \showarticletitle{Mel-scaled discrete wavelet coefficients for speech
  recognition}. In \bibinfo{booktitle}{\emph{ICASSP}}.
\newblock


\bibitem[\protect\citeauthoryear{He, Zhang, Ren, and Sun}{He
  et~al\mbox{.}}{2015}]%
        {ResNet}
\bibfield{author}{\bibinfo{person}{Kaiming He}, \bibinfo{person}{Xiangyu
  Zhang}, \bibinfo{person}{Shaoqing Ren}, {and} \bibinfo{person}{Jian Sun}.}
  \bibinfo{year}{2015}\natexlab{}.
\newblock \showarticletitle{Deep Residual Learning for Image Recognition}. In
  \bibinfo{booktitle}{\emph{CVPR}}.
\newblock


\bibitem[\protect\citeauthoryear{Heilbron, Escorcia, Ghanem, and
  Niebles}{Heilbron et~al\mbox{.}}{2015}]%
        {activitynet}
\bibfield{author}{\bibinfo{person}{Fabian~Caba Heilbron},
  \bibinfo{person}{Victor Escorcia}, \bibinfo{person}{Bernard Ghanem}, {and}
  \bibinfo{person}{Juan~Carlos Niebles}.} \bibinfo{year}{2015}\natexlab{}.
\newblock \showarticletitle{ActivityNet: A Large-Scale Video Benchmark for
  Human Activity Understanding}. In \bibinfo{booktitle}{\emph{CVPR}}.
\newblock


\bibitem[\protect\citeauthoryear{Ioffe and Szegedy}{Ioffe and Szegedy}{2015}]%
        {InceptionV2}
\bibfield{author}{\bibinfo{person}{Sergey Ioffe} {and}
  \bibinfo{person}{Christian Szegedy}.} \bibinfo{year}{2015}\natexlab{}.
\newblock \showarticletitle{Batch Normalization: Accelerating Deep Network
  Training by Reducing Internal Covariate Shift}. In
  \bibinfo{booktitle}{\emph{ICML}}.
\newblock


\bibitem[\protect\citeauthoryear{Jiang, Dai, Xue, Liu, and Ngo}{Jiang
  et~al\mbox{.}}{2012}]%
        {jiang2012trajectory}
\bibfield{author}{\bibinfo{person}{Yugang Jiang}, \bibinfo{person}{Qi Dai},
  \bibinfo{person}{Xiangyang Xue}, \bibinfo{person}{Wei Liu}, {and}
  \bibinfo{person}{Chong-Wah Ngo}.} \bibinfo{year}{2012}\natexlab{}.
\newblock \showarticletitle{Trajectory-based modeling of human actions with
  motion reference points}. In \bibinfo{booktitle}{\emph{ECCV}}.
\newblock


\bibitem[\protect\citeauthoryear{Jiang, Liu, Zamir, Laptev, Piccardi, Shah, and
  Sukthankar}{Jiang et~al\mbox{.}}{2013a}]%
        {THUMOS15}
\bibfield{author}{\bibinfo{person}{Yugang Jiang}, \bibinfo{person}{Jingen Liu},
  \bibinfo{person}{Amir~Roshan Zamir}, \bibinfo{person}{Ivan Laptev},
  \bibinfo{person}{Massimo Piccardi}, \bibinfo{person}{Mubarak Shah}, {and}
  \bibinfo{person}{Rahul Sukthankar}.} \bibinfo{year}{2013}\natexlab{a}.
\newblock \bibinfo{title}{THUMOS Challenge: Action Recognition with a Large
  Number of Classes}.
\newblock
  \bibinfo{howpublished}{\url{http://crcv.ucf.edu/ICCV13-Action-Workshop/}}.
  (\bibinfo{year}{2013}).
\newblock


\bibitem[\protect\citeauthoryear{Jiang, Wang, Feng, Xue, Zheng, and Yang}{Jiang
  et~al\mbox{.}}{2013b}]%
        {Jiang}
\bibfield{author}{\bibinfo{person}{Yugang Jiang}, \bibinfo{person}{Yanran
  Wang}, \bibinfo{person}{Rui Feng}, \bibinfo{person}{Xiangyang Xue},
  \bibinfo{person}{Yingbin Zheng}, {and} \bibinfo{person}{Hanfang Yang}.}
  \bibinfo{year}{2013}\natexlab{b}.
\newblock \showarticletitle{Understanding and Predicting Interestingness of
  Videos}. In \bibinfo{booktitle}{\emph{AAAI}}.
\newblock


\bibitem[\protect\citeauthoryear{Jiang, Wu, Wang, Xue, and Chang}{Jiang
  et~al\mbox{.}}{2015}]%
        {FCVID}
\bibfield{author}{\bibinfo{person}{Yugang Jiang}, \bibinfo{person}{Zuxuan Wu},
  \bibinfo{person}{Jun Wang}, \bibinfo{person}{Xiangyang Xue}, {and}
  \bibinfo{person}{Shih-Fu Chang}.} \bibinfo{year}{2015}\natexlab{}.
\newblock \showarticletitle{Exploiting Feature and Class Relationships in Video
  Categorization with Regularized Deep Neural Networks}.
\newblock \bibinfo{journal}{\emph{arXiv preprint arXiv:1502.07209}}
  (\bibinfo{year}{2015}).
\newblock


\bibitem[\protect\citeauthoryear{Karpathy, Toderici, Shetty, Leung, Sukthankar,
  and Fei-Fei}{Karpathy et~al\mbox{.}}{2014}]%
        {Sports-1M}
\bibfield{author}{\bibinfo{person}{Andrej Karpathy}, \bibinfo{person}{George
  Toderici}, \bibinfo{person}{Sanketh Shetty}, \bibinfo{person}{Thomas Leung},
  \bibinfo{person}{Rahul Sukthankar}, {and} \bibinfo{person}{Li Fei-Fei}.}
  \bibinfo{year}{2014}\natexlab{}.
\newblock \showarticletitle{Large-scale Video Classification with Convolutional
  Neural Networks}. In \bibinfo{booktitle}{\emph{CVPR}}.
\newblock


\bibitem[\protect\citeauthoryear{Khosla, Das~Sarma, and Hamid}{Khosla
  et~al\mbox{.}}{2014}]%
        {khosla2014makes}
\bibfield{author}{\bibinfo{person}{Aditya Khosla}, \bibinfo{person}{Atish
  Das~Sarma}, {and} \bibinfo{person}{Raffay Hamid}.}
  \bibinfo{year}{2014}\natexlab{}.
\newblock \showarticletitle{What makes an image popular?}. In
  \bibinfo{booktitle}{\emph{WWW}}.
\newblock


\bibitem[\protect\citeauthoryear{Kuehne, Jhuang, Garrote, Poggio, and
  Serre}{Kuehne et~al\mbox{.}}{2011}]%
        {HMDB}
\bibfield{author}{\bibinfo{person}{H. Kuehne}, \bibinfo{person}{H. Jhuang},
  \bibinfo{person}{E. Garrote}, \bibinfo{person}{T. Poggio}, {and}
  \bibinfo{person}{T. Serre}.} \bibinfo{year}{2011}\natexlab{}.
\newblock \showarticletitle{HMDB: a large video database for human motion
  recognition}. In \bibinfo{booktitle}{\emph{ICCV}}.
\newblock


\bibitem[\protect\citeauthoryear{Laptev and Ivan}{Laptev and Ivan}{2005}]%
        {KTH}
\bibfield{author}{\bibinfo{person}{Laptev} {and} \bibinfo{person}{Ivan}.}
  \bibinfo{year}{2005}\natexlab{}.
\newblock \showarticletitle{On space-time interest points}.
\newblock \bibinfo{journal}{\emph{International journal of computer vision}}
  \bibinfo{volume}{64}, \bibinfo{number}{2-3} (\bibinfo{year}{2005}),
  \bibinfo{pages}{107--123}.
\newblock


\bibitem[\protect\citeauthoryear{Li, Song, Cao, Tetreault, Goldberg, Jaimes,
  and Luo}{Li et~al\mbox{.}}{2016}]%
        {li2016tgif}
\bibfield{author}{\bibinfo{person}{Yuncheng Li}, \bibinfo{person}{Yale Song},
  \bibinfo{person}{Liangliang Cao}, \bibinfo{person}{Joel Tetreault},
  \bibinfo{person}{Larry Goldberg}, \bibinfo{person}{Alejandro Jaimes}, {and}
  \bibinfo{person}{Jiebo Luo}.} \bibinfo{year}{2016}\natexlab{}.
\newblock \showarticletitle{TGIF: A new dataset and benchmark on animated GIF
  description}. In \bibinfo{booktitle}{\emph{CVPR}}.
\newblock


\bibitem[\protect\citeauthoryear{Liu, Niu, and Gleicher}{Liu
  et~al\mbox{.}}{2009}]%
        {Liu2009}
\bibfield{author}{\bibinfo{person}{Feng Liu}, \bibinfo{person}{Yuzhen Niu},
  {and} \bibinfo{person}{Michael Gleicher}.} \bibinfo{year}{2009}\natexlab{}.
\newblock \showarticletitle{Using Web Photos for Measuring Video Frame
  Interestingness}. In \bibinfo{booktitle}{\emph{IJCAI}}.
\newblock


\bibitem[\protect\citeauthoryear{Liu, Li, and Tang}{Liu et~al\mbox{.}}{2006}]%
        {liu2006spatio}
\bibfield{author}{\bibinfo{person}{Wei Liu}, \bibinfo{person}{Zhifeng Li},
  {and} \bibinfo{person}{Xiaoou Tang}.} \bibinfo{year}{2006}\natexlab{}.
\newblock \showarticletitle{Spatio-temporal embedding for statistical face
  recognition from video}. In \bibinfo{booktitle}{\emph{ECCV}}.
\newblock


\bibitem[\protect\citeauthoryear{Marsza{\l}ek, Laptev, and Schmid}{Marsza{\l}ek
  et~al\mbox{.}}{2009}]%
        {Hollywood2}
\bibfield{author}{\bibinfo{person}{Marcin Marsza{\l}ek}, \bibinfo{person}{Ivan
  Laptev}, {and} \bibinfo{person}{Cordelia Schmid}.}
  \bibinfo{year}{2009}\natexlab{}.
\newblock \showarticletitle{Actions in Context}. In
  \bibinfo{booktitle}{\emph{CVPR}}.
\newblock


\bibitem[\protect\citeauthoryear{Messer, Matas, Kittler, Luettin, and
  Maitre}{Messer et~al\mbox{.}}{1999}]%
        {Messer-avbpa99}
\bibfield{author}{\bibinfo{person}{K Messer}, \bibinfo{person}{J Matas},
  \bibinfo{person}{J Kittler}, \bibinfo{person}{J Luettin}, {and}
  \bibinfo{person}{G Maitre}.} \bibinfo{year}{1999}\natexlab{}.
\newblock \showarticletitle{XM2VTSDB: The Extended M2VTS Database}. In
  \bibinfo{booktitle}{\emph{Second International Conference on Audio and
  Video-based Biometric Person Authentication}}.
\newblock


\bibitem[\protect\citeauthoryear{Miech, Laptev, and Sivic}{Miech
  et~al\mbox{.}}{2017}]%
        {miech2017learnable}
\bibfield{author}{\bibinfo{person}{Antoine Miech}, \bibinfo{person}{Ivan
  Laptev}, {and} \bibinfo{person}{Josef Sivic}.}
  \bibinfo{year}{2017}\natexlab{}.
\newblock \showarticletitle{Learnable pooling with Context Gating for video
  classification}.
\newblock \bibinfo{journal}{\emph{arXiv preprint arXiv:1706.06905}}
  (\bibinfo{year}{2017}).
\newblock


\bibitem[\protect\citeauthoryear{Oord, Dieleman, Zen, Simonyan, Vinyals,
  Graves, Kalchbrenner, Senior, and Kavukcuoglu}{Oord et~al\mbox{.}}{2016}]%
        {wavenet}
\bibfield{author}{\bibinfo{person}{Aaron van~den Oord}, \bibinfo{person}{Sander
  Dieleman}, \bibinfo{person}{Heiga Zen}, \bibinfo{person}{Karen Simonyan},
  \bibinfo{person}{Oriol Vinyals}, \bibinfo{person}{Alex Graves},
  \bibinfo{person}{Nal Kalchbrenner}, \bibinfo{person}{Andrew Senior}, {and}
  \bibinfo{person}{Koray Kavukcuoglu}.} \bibinfo{year}{2016}\natexlab{}.
\newblock \showarticletitle{Wavenet: A generative model for raw audio}.
\newblock \bibinfo{journal}{\emph{arXiv preprint arXiv:1609.03499}}
  (\bibinfo{year}{2016}).
\newblock


\bibitem[\protect\citeauthoryear{Pinto, Almeida, and Gon\c{c}alves}{Pinto
  et~al\mbox{.}}{2013}]%
        {Pinto:2013:UEV:2433396.2433443}
\bibfield{author}{\bibinfo{person}{Henrique Pinto}, \bibinfo{person}{Jussara~M.
  Almeida}, {and} \bibinfo{person}{Marcos~A. Gon\c{c}alves}.}
  \bibinfo{year}{2013}\natexlab{}.
\newblock \showarticletitle{Using Early View Patterns to Predict the Popularity
  of Youtube Videos}. In \bibinfo{booktitle}{\emph{WSDM}}.
\newblock


\bibitem[\protect\citeauthoryear{Rizoiu, Xie, Sanner, Cebrian, Yu, and
  Van~Hentenryck}{Rizoiu et~al\mbox{.}}{2017}]%
        {Rizoiu:2017:EHH:3038912.3052650}
\bibfield{author}{\bibinfo{person}{Marian-Andrei Rizoiu},
  \bibinfo{person}{Lexing Xie}, \bibinfo{person}{Scott Sanner},
  \bibinfo{person}{Manuel Cebrian}, \bibinfo{person}{Honglin Yu}, {and}
  \bibinfo{person}{Pascal Van~Hentenryck}.} \bibinfo{year}{2017}\natexlab{}.
\newblock \showarticletitle{Expecting to Be HIP: Hawkes Intensity Processes for
  Social Media Popularity}. In \bibinfo{booktitle}{\emph{WWW}}.
\newblock


\bibitem[\protect\citeauthoryear{Shen, Demarty, and Duong}{Shen
  et~al\mbox{.}}{2016}]%
        {MediaEval}
\bibfield{author}{\bibinfo{person}{Yuesong Shen},
  \bibinfo{person}{Claire-H{\'e}l{\`e}ne Demarty}, {and} \bibinfo{person}{Ngoc
  Q.~K. Duong}.} \bibinfo{year}{2016}\natexlab{}.
\newblock \showarticletitle{Technicolor@MediaEval 2016 Predicting Media
  Interestingness Task}. In \bibinfo{booktitle}{\emph{MediaEval}}.
\newblock


\bibitem[\protect\citeauthoryear{Simonyan and Zisserman}{Simonyan and
  Zisserman}{2014}]%
        {vgg}
\bibfield{author}{\bibinfo{person}{Karen Simonyan} {and}
  \bibinfo{person}{Andrew Zisserman}.} \bibinfo{year}{2014}\natexlab{}.
\newblock \showarticletitle{Very Deep Convolutional Networks for Large-Scale
  Image Recognition}.
\newblock \bibinfo{journal}{\emph{arXiv preprint arXiv:1409.1556}}
  (\bibinfo{year}{2014}).
\newblock


\bibitem[\protect\citeauthoryear{Soomro, Zamir, and Shah}{Soomro
  et~al\mbox{.}}{2012}]%
        {UCF-101}
\bibfield{author}{\bibinfo{person}{Khurram Soomro},
  \bibinfo{person}{Amir~Roshan Zamir}, {and} \bibinfo{person}{Mubarak Shah}.}
  \bibinfo{year}{2012}\natexlab{}.
\newblock \showarticletitle{UCF101: A Dataset of 101 Human Actions Classes From
  Videos in The Wild}. In \bibinfo{booktitle}{\emph{arXiv:1212.0402}}.
\newblock


\bibitem[\protect\citeauthoryear{Szab{\'{o}} and Huberman}{Szab{\'{o}} and
  Huberman}{2010}]%
        {DBLP:journals/cacm/SzaboH10}
\bibfield{author}{\bibinfo{person}{G{\'{a}}bor Szab{\'{o}}} {and}
  \bibinfo{person}{Bernardo~A. Huberman}.} \bibinfo{year}{2010}\natexlab{}.
\newblock \showarticletitle{Predicting the popularity of online content}.
\newblock \bibinfo{journal}{\emph{Commun. {ACM}}} \bibinfo{volume}{53},
  \bibinfo{number}{8} (\bibinfo{year}{2010}), \bibinfo{pages}{80--88}.
\newblock


\bibitem[\protect\citeauthoryear{Szegedy, Ioffe, Vanhoucke, and Alemi}{Szegedy
  et~al\mbox{.}}{2016a}]%
        {InceptionV4}
\bibfield{author}{\bibinfo{person}{Christian Szegedy}, \bibinfo{person}{Sergey
  Ioffe}, \bibinfo{person}{Vincent Vanhoucke}, {and} \bibinfo{person}{Alex~A.
  Alemi}.} \bibinfo{year}{2016}\natexlab{a}.
\newblock \showarticletitle{Inception-v4, Inception-ResNet and the Impact of
  Residual Connections on Learning}. In \bibinfo{booktitle}{\emph{ICLR
  Workshop}}.
\newblock


\bibitem[\protect\citeauthoryear{Szegedy, Liu, Jia, Sermanet, Reed, Anguelov,
  Erhan, Vanhoucke, and Rabinovich}{Szegedy et~al\mbox{.}}{2015}]%
        {InceptionV1}
\bibfield{author}{\bibinfo{person}{Christian Szegedy}, \bibinfo{person}{Wei
  Liu}, \bibinfo{person}{Yangqing Jia}, \bibinfo{person}{Pierre Sermanet},
  \bibinfo{person}{Scott Reed}, \bibinfo{person}{Dragomir Anguelov},
  \bibinfo{person}{Dumitru Erhan}, \bibinfo{person}{Vincent Vanhoucke}, {and}
  \bibinfo{person}{Andrew Rabinovich}.} \bibinfo{year}{2015}\natexlab{}.
\newblock \showarticletitle{Going Deeper with Convolutions}. In
  \bibinfo{booktitle}{\emph{CVPR}}.
\newblock


\bibitem[\protect\citeauthoryear{Szegedy, Vanhoucke, Ioffe, Shlens, and
  Wojna}{Szegedy et~al\mbox{.}}{2016b}]%
        {InceptionV3}
\bibfield{author}{\bibinfo{person}{Christian Szegedy}, \bibinfo{person}{Vincent
  Vanhoucke}, \bibinfo{person}{Sergey Ioffe}, \bibinfo{person}{Jonathon
  Shlens}, {and} \bibinfo{person}{Zbigniew Wojna}.}
  \bibinfo{year}{2016}\natexlab{b}.
\newblock \showarticletitle{Rethinking the Inception Architecture for Computer
  Vision}. In \bibinfo{booktitle}{\emph{CVPR}}.
\newblock


\bibitem[\protect\citeauthoryear{Thomee, Shamma, Friedland, Elizalde, Ni,
  Poland, Borth, and Li}{Thomee et~al\mbox{.}}{2016}]%
        {YFCC}
\bibfield{author}{\bibinfo{person}{Bart Thomee}, \bibinfo{person}{David~A.
  Shamma}, \bibinfo{person}{Gerald Friedland}, \bibinfo{person}{Benjamin
  Elizalde}, \bibinfo{person}{Karl Ni}, \bibinfo{person}{Douglas Poland},
  \bibinfo{person}{Damian Borth}, {and} \bibinfo{person}{Li-Jia Li}.}
  \bibinfo{year}{2016}\natexlab{}.
\newblock \showarticletitle{YFCC100M: The New Data in Multimedia Research}.
\newblock \bibinfo{journal}{\emph{Commun. ACM}} \bibinfo{volume}{59},
  \bibinfo{number}{2} (\bibinfo{date}{Jan.} \bibinfo{year}{2016}),
  \bibinfo{pages}{64--73}.
\newblock
\showISSN{0001-0782}


\bibitem[\protect\citeauthoryear{Tzanetakis and Cook}{Tzanetakis and
  Cook}{2002}]%
        {tzanetakis2002musical}
\bibfield{author}{\bibinfo{person}{George Tzanetakis} {and}
  \bibinfo{person}{Perry Cook}.} \bibinfo{year}{2002}\natexlab{}.
\newblock \showarticletitle{Musical genre classification of audio signals}.
\newblock \bibinfo{journal}{\emph{IEEE Transactions on speech and audio
  processing}} \bibinfo{volume}{10}, \bibinfo{number}{5}
  (\bibinfo{year}{2002}), \bibinfo{pages}{293--302}.
\newblock


\bibitem[\protect\citeauthoryear{Wu, Zhong, Tan, Horner, and Yang}{Wu
  et~al\mbox{.}}{2014}]%
        {DBLP:conf/kdd/WuZTH014}
\bibfield{author}{\bibinfo{person}{Bin Wu}, \bibinfo{person}{Erheng Zhong},
  \bibinfo{person}{Ben Tan}, \bibinfo{person}{Andrew Horner}, {and}
  \bibinfo{person}{Qiang Yang}.} \bibinfo{year}{2014}\natexlab{}.
\newblock \showarticletitle{Crowdsourced time-sync video tagging using temporal
  and personalized topic modeling}. In \bibinfo{booktitle}{\emph{SIGKDD}}.
\newblock


\bibitem[\protect\citeauthoryear{Yoon and Pavlovic}{Yoon and Pavlovic}{2014}]%
        {SentimentFlow}
\bibfield{author}{\bibinfo{person}{Sejong Yoon} {and} \bibinfo{person}{Vladimir
  Pavlovic}.} \bibinfo{year}{2014}\natexlab{}.
\newblock \showarticletitle{Sentiment Flow for Video Interestingness
  Prediction}. In \bibinfo{booktitle}{\emph{Proceedings of the 1st ACM
  International Workshop on Human Centered Event Understanding from
  Multimedia}} \emph{(\bibinfo{series}{HuEvent '14})}.
\newblock


\bibitem[\protect\citeauthoryear{Zoph, Vasudevan, Shlens, and Le}{Zoph
  et~al\mbox{.}}{2017}]%
        {nasnet}
\bibfield{author}{\bibinfo{person}{Barret Zoph}, \bibinfo{person}{Vijay
  Vasudevan}, \bibinfo{person}{Jonathon Shlens}, {and} \bibinfo{person}{Quoc~V.
  Le}.} \bibinfo{year}{2017}\natexlab{}.
\newblock \showarticletitle{Learning transferable architectures for scalable
  image recognition}.
\newblock \bibinfo{journal}{\emph{arXiv preprint arXiv:1707.07012}}
  (\bibinfo{year}{2017}).
\newblock


\end{thebibliography}

\end{document}